\documentclass[letterpaper, 10 pt, conference]{ieeeconf}  

\IEEEoverridecommandlockouts                              

\usepackage{graphicx}
\usepackage{subcaption}
\usepackage{booktabs} 
\usepackage{tabulary}
\usepackage{setspace}
\usepackage{bm}
\usepackage{amsmath,amssymb, amsthm} 
\usepackage{hyperref}
\hypersetup{
    hidelinks,
    colorlinks=true,
    allcolors=blue
}
\usepackage{color}
\usepackage{cite}
\usepackage{multicol}
\usepackage{algorithm}
\usepackage{algorithmic}
\usepackage{tabularx}
\usepackage{todonotes}

\theoremstyle{definition}
\newtheorem{definition}{Definition}

\theoremstyle{plain}

\title{\LARGE \bf
Asynchronous, Option-Based Multi-Agent Policy Gradient: A Conditional Reasoning Approach
}

\author{Xubo Lyu$^{1,2}$, Amin Banitalebi-Dehkordi$^{2}$, Mo Chen$^{1}$, Yong Zhang$^{2}$
\thanks{$^{1}$ School of Computing Science, Simon Fraser University, BC, Canada}%
\thanks{$^{2}$ Huawei Technologies Canada Co., Ltd. Burnaby, BC, Canada.}
}

\begin{document}

\maketitle
\thispagestyle{empty}
\pagestyle{empty}

\begin{abstract}

Cooperative multi-agent problems often require coordination between agents, which can be achieved through a centralized policy that considers the global state.
Multi-agent policy gradient (MAPG) methods are commonly used to learn such policies, but they are often limited to problems with low-level action spaces. 
In complex problems with large state and action spaces, it is advantageous to extend MAPG methods to use higher-level actions, also known as options, to improve the policy search efficiency.
However, multi-robot option executions are often asynchronous, that is, agents may select and complete their options at different time steps. This makes it difficult for MAPG methods to derive a centralized policy and evaluate its gradient, as centralized policy always select new options at the same time. In this work, we propose a novel, conditional reasoning approach to address this problem and demonstrate its effectiveness on representative option-based multi-agent cooperative tasks through empirical validation. Find code and videos at: \href{https://sites.google.com/view/mahrlsupp/}{https://sites.google.com/view/mahrlsupp/}

\end{abstract}

\section{introduction}
Cooperative multi-agent problems is common in real-world. For example, a team of warehouse agents must coordinate their behaviours for efficient cargo handling, or a group of robots to achieve a common goal in a team-based game. In these cases, agents' actions and outcomes affect each other, thus a centralized controller is often needed to learn a policy that takes the joint observation of all agents as input and outputs a joint action for all agents \cite{gupta2017cooperative, fu2022revisiting, xuan2002multi}. 

To efficiently learn a cooperative, centralized policy, multi-agent policy gradient (MAPG) methods have been widely applied and have many successful applications \cite{rashid2018qmix, foerster2018counterfactual, yu2021surprising, wiederer2022anomaly, wang2022influencing, wang2020model}. One key feature of MAPG methods is that they normally require robots to perform low-level actions at a low-level time scale: an action only lasts one time step and multiple agents choose their actions synchronously at every time step.
Although low-level actions can be useful, they are not always practical for solving complex problems especially those with large state and action space. This is because relying solely on  low-level actions may lead to a combinatorial explosion, where there are too many possible actions to search, making it difficult to find an optimal policy. 

\begin{figure}[!htp]
\centering
\includegraphics[width=0.45\textwidth]{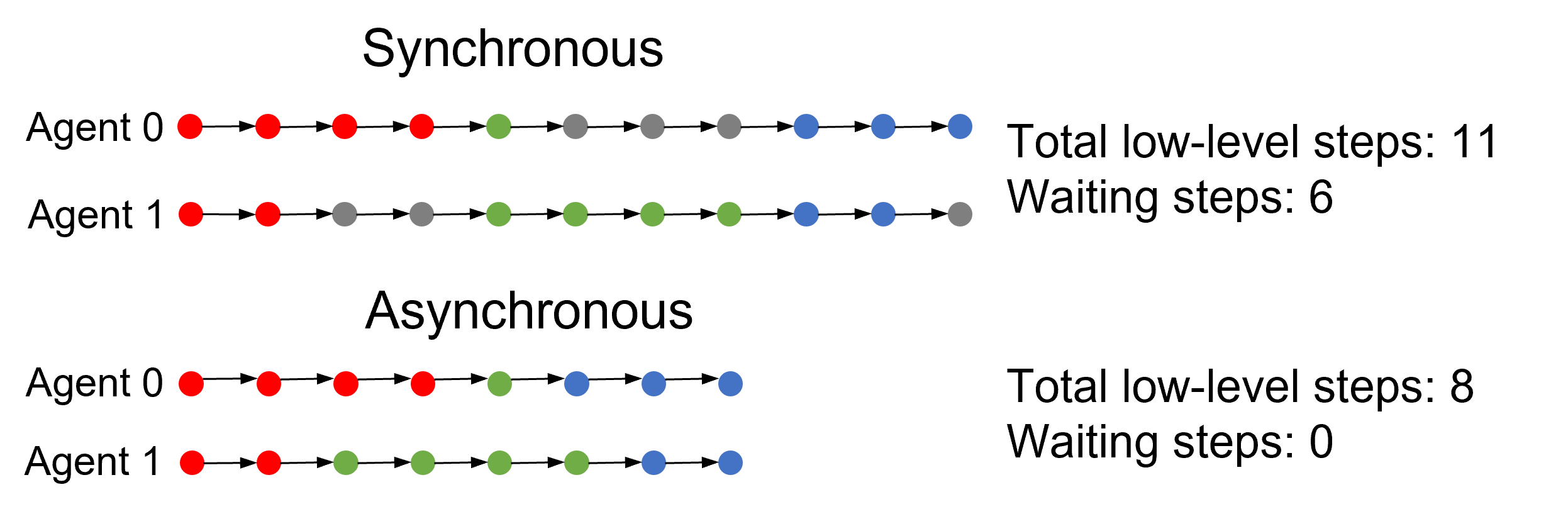}
\caption{\small An illustration of synchronous and asynchronous option execution. Each node refers to a state. Each arrow refers to a low-level action. A sequence of nodes in red, green or blue refers to executing one option. The gray nodes are waiting states for option synchronization. Asynchronous strategy is more efficient and   requires much less low-level waiting steps.
}
\label{fig:intro-concept}
\end{figure}

To improve policy search efficiency, hierarchical reinforcement learning (HRL) has been developed and widely adopted \cite{makar2001hierarchical, kulkarni2016hierarchical, sutton1999between, frans2018meta, xu2021hierarchical}. HRL breaks down complex tasks into smaller subtasks to reduce the problem complexity, making it easier to search for policies. This is especially helpful for long-term reasoning tasks with sparse reward feedback.
One typical approach of HRL is the options framework \cite{sutton1999between}, which employs options to optimize policies. Options are higher-level combinations of primitive actions that enable agents to perform actions over extended duration. The use of options effectively reduces search space size and improves policy learning efficiency and generalisability \cite{yang2020hierarchical, voigt2021multi, stone2005reinforcement}.
In order to effectively address complex cooperative problems, it is advantageous to employ option framework into MAPG methods for centralized learning. However, a significant challenge arises from the inconsistency between asynchronous option execution and centralized decision-making. This inconsistency results from the fact that options have varying low-level time lengths, leading to their selection and completion across all agents occurring at different time steps, that is, "asynchronicity" of multi-agent option execution.
While preserving asynchronicity can enable a realistic and efficient option execution strategy, as illustrated in Fig.~\ref{fig:intro-concept}, it also poses an obstacle to MAPG methods in determining when and how to optimize a joint control policy and sample new options from it for all agents. Specifically, under asynchronicity, the joint option sampled from a centralized policy may not be fully executed. For example, at a given low-level time step, only a subset of agents may need to choose new options, while others do not. Therefore, it is unclear how to properly evaluate policy gradients for policy training.

Several attempts \cite{xiao2022asynchronous, yang2020hierarchical, han2019multi, amato2019modeling, pmlr-v100-xiao20a} have been made to address the aforementioned challenge, but they have limitations. Specifically, these attempts can be classified into two categories: synchronous and asynchronous approaches. Synchronous approaches \cite{yang2020hierarchical, han2019multi} force all agents' options to be synchronized at certain low-level time steps. For example, they either use $\eta_\text{any}$ strategy that interrupts unfinished options when \textit{any} one of the agents finish its option, or use  $\eta_\text{all}$ strategy that waits until \textit{all} agents' options to finish. Then the standard MAPG method can be applied to make centralized decisions as if they are working on synchronous actions with an "extended step".
However, this type of approaches destroy asynchronicity, and  eliminates the option completeness as well as its benefits of low-level temporal abstraction.
In contrast, Christopher et al. \cite{amato2019modeling, pmlr-v100-xiao20a} proposed an alternative, asynchronous solution. It allows the use of  $\eta_\text{continue}$ strategy:  a subset of agents choose
new options while the others continue their ongoing options. They achieve this by  aligning multi-agent options via a specially-designed sequence filtering process. This method preserves asynchronicity but is only compatible with value-based algorithms (e.g., Deep Q-Network (DQN)\cite{mnih2013playing}), not policy gradient algorithms.

In this work, we propose a novel, conditional reasoning approach to  enable centralized learning of MAPG over asynchronous options.
%
To achieve this, we first introduce a special type of trajectory named ``option-level joint trajectory'', which combines multi-agent independent trajectories into one single joint trajectory. This trajectory allows agents to reason and plan at a option level while preserve efficient, asynchronous multi-robot option execution. 
To generate such trajectory for policy training, we formalize a conditional centralized policy that only selects a subset of options for those agents who require new options, but conditions its policy distribution on the currently-executing options. 
In this way, we address the inconsistency between option asynchronicity and centralized decision-making. As a result, our method can  seamlessly adapt MAPG from action-wise to option-wise while fitting it into centralized learning.
Our method is simple and can be easily implemented as an add-on to standard, action-based policy gradient algorithms, such as MAPPO \cite{yu2021surprising}, to extend its use over asynchronous options. We validate the effectiveness of our method on two option-based multi-agent cooperative learning tasks.


\section{Background}
\subsection{Problem Formulation}
Following option framework \cite{sutton1999between}, we formulate a two-level, multi-agent HRL problem for cooperatively training ${N}$ agents. In the higher level, at the option step $t$ (its corresponding low-level time step $k$), an option-based joint policy $\pi(o_t|z_t) \in [0,1]$ needs to be trained to determine a joint option $o_t$  based on a joint observation $z_t$ (under true state $s_t$) for ${N}$ agents. $o_t = (o_t^0, o_t^1, ... o_t^{N-1})$ and $z_t = (z_t^0, z_t^1, ... z_t^{N-1})$ are collections of each agent $i$'s option and observation, where $i$ is indexed by $i=\{0, 1, 2, ..., N-1\}$. In the lower level, each agent observes the local observation $z_k^i$ under the true state $s_k^i$, and takes an action $a_k^i$ which is determined by the inner policy $\pi_{o_t^i}(a_k^i|z_k^i)$ of current-executing option  $o_t^{i}$. After all agents take their actions, the environment transits from $s_{k}$ to the next state $s_{k+1}$ according to $P(\cdot|s_k, a_k)$ and achieve a joint reward $R(s_k, a_k)$.
In our setup, the transition function $P(\cdot|s_k, a_k)$ is based on low-level time step, and is  determined by the simulation. 

Our objective is to find the higher-level, option-based policy $\pi(o|z)$, rather the inner policy $\pi_o$ over lower-level action space.  Therefore, we assume the inner, action-based policies $\pi_{o}$ of available options are given or pre-defined that are appropriate for potential task completion. This is to say, we are not learning $\pi_{o}$ in the low-level.

\begin{figure*}[!htp]
\centering
\includegraphics[width=0.9\textwidth]{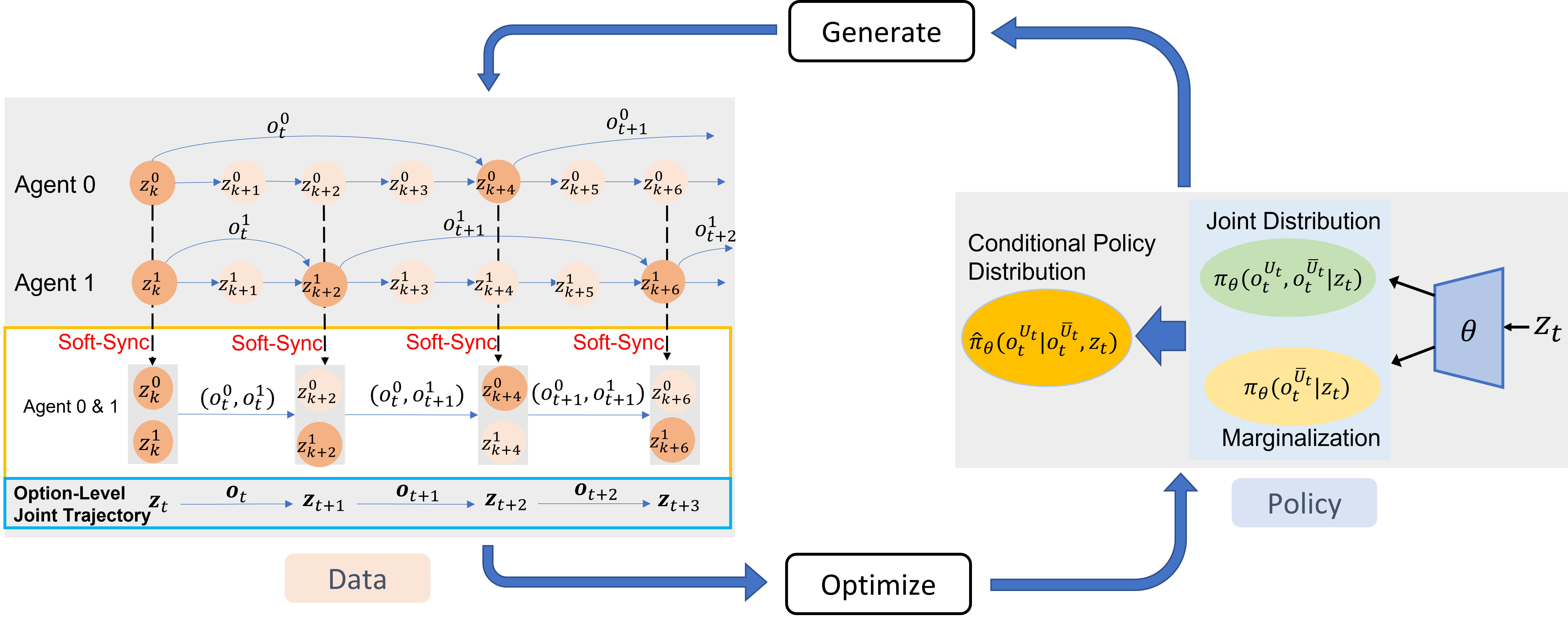}
\caption{\small The structure of our method. \textbf{Left:} an example of option-level joint trajectory where two agents take their options asynchronously. Observations  ${z}_t,{z}_{t+1},... $ are joint observations of all agents at option step $t, t+1, ...$ when at least one agent's option is terminated. Option ${o}_t, 
{o}_{t+1}, ...$ are joint options at each option step. Batches of these trajectories are used as training data to optimize the conditional centralized policy. \textbf{Right:} The formation of centralized conditional policy $\hat{\pi}_{\theta}$ where $\theta$ is the policy parameter. This policy is used to generate required joint option for a subset of agents $U_t$  to form the option-level trajectories. }
\label{fig:method}
\end{figure*}

\subsection{Centralized MAPG over Asynchronous Options}
In centralized learning, asynchronous option execution presents a problem for policy gradient optimization. To understand the problem, let us consider Generalized Advantage Estimation (GAE)-based  policy gradient \cite{schulman2017proximal} shown in Eq.~\eqref{eq:gae_actor_critic}.
It describes a centralized controller selecting the joint action $a_k$ for all agents at the low-level time scale indexed by $k$.  It calculates the policy gradient of the RL objective $J(\pi_{\theta})$ over a batch of trajectories $\tau$. $G_{k}^{\lambda}$ estimates the Temporal-Difference TD($\lambda$) \cite{sutton1988learning} return   and $V_{\pi_{\theta}}$ estimates the value function which is used as a baseline to lower the gradient variance. Our question is: \textit{how to adapt Eq.~\eqref{eq:gae_actor_critic}, which is based on actions, to the use of options, and apply it to learn a centralized policy?  }
\begin{equation} 
\label{eq:gae_actor_critic}
\scriptsize
    \nabla_{\theta} J\left(\pi_{\theta}\right)=\underset{\tau \sim \pi_{\theta}}{\mathrm{E}}\left[\sum_{k=0}^{K} \nabla_{\theta} \log \pi_{\theta}\left(a_{k} \mid z_{k}\right)\left(G_{k}^{\lambda} - V_{\pi_{\theta}}(z_k)\right)\right].
\end{equation} 

One trivial way is to simply substitute action-based policy $\pi(a_k|z_k)$ with option-based policy $\pi(o_t|z_t)$ and  view an option as an action with ``one extended step". In this way, Eq.~\eqref{eq:gae_actor_critic} still holds, and can be converted to base on option time $t$  defined by the option termination, rather action time $k$. However, this only works when multi-agent options are forced to be synchronized (e.g. $\eta_\text{any}$ or $\eta_\text{all}$) such that a joint policy can always select new options for all agents at the same time. But it is not applicable to many realistic situations where the asynchronous option execution $\eta_\text{continue}$ is preferred. In fact, the calculation for policy term $\pi(o_t|z_t)$ remains a challenge when $\eta_\text{continue}$ strategy is used, as there is no ``synchronous timing" for a joint option selection any more. Therefore, we present a novel solution to address it in the next section.

\section{Conditional Centralized MAPG over Asynchronous Options}
\label{sec:method}

\subsection{Option-Level Joint Trajectory}
\label{sec:methodA}
To train a multi-agent option-based policy in centralized way, we require a collection of agents' trajectories as training data. Unlike action-level trajectory that is often defined at a sequence of low-level time steps, here we introduce a special option-level trajectory in Fig.~\ref{fig:method}. To form it, we determine specific low-level timings at when we collect {joint} states and observations, {joint} options and the shared reward for all agents. Those timings are when \textit{at least one agent} terminates its previous option and selects a new option, and they will serve as option time steps in the trajectory. We refer to this process as "soft-sync" because it synchronizes multi-agent state transitions and option selections in a soft way --- it does not actually interrupt agents' ongoing options.

In contrast to strategies like $\eta_\text{any}$ or $\eta_\text{all}$ that actually interrupts robot-environment interactions to have forced synchronization,  ``soft-sync'' allows the formation of an ``interruption-free'' trajectory over asynchronous options, and allows the use of a centralized policy to jointly select options. 
The trajectory is formally described  by Definition~\ref{definition_2}.

\begin{definition}
\label{definition_2}
We denote $\tau_o$ as an option-level, joint trajectory for $N$ agents. $\tau_o=(z_{t}, o_{t}, r_{t}, z_{t+1}, o_{t+1}, r_{t+1}, ..., z_{t+T})$ where $t, t+1, ..., t+T$ are consecutive option steps when ``soft-sync'' happens. In particular, $z$ and $o$ are joint observation and joint option: $z_t = (z_t^0, z_t^1, ..., z_t^{N-1})$ and $o_t = (o_t^0, o_t^1, ... o_t^{N-1})$. $r_t$ is shared step reward at time $t$. 
\end{definition}

\subsection{Conditional Centralized Policy} 
\label{methodB}

To achieve centralized MAPG over asynchronous options, we not only requires option-level joint trajectories $\tau_o$, but also a special form of policy to generate these trajectories and get updated from them. 
This is because many policy gradient methods \cite{schulman2017proximal, schulman2015trust, yu2021surprising} update policy in an \textit{on-policy} way: the policy being updated is also used to generate sequential data for training itself.
%

Although sampling new options for all agents from a joint policy $\pi(o_t|z_t)$ at each time step $t$ is the easiest approach, it is infeasible. 
The reason is that only a subset of agents need to select new options at each option step $t$ of an option-level joint trajectory, while the others need to continue with their ongoing options to maintain "interruption-free" asynchronicity. 
For example, in Fig.~\ref{fig:method}, at option step $t+1$, agent 1 requires a new option, but agent 0 does not and  continue with its ongoing option. 
%
%
Thus, a policy would be desirable if it can generate joint options at each option step only for those agents who need new options while not affecting the other agents with on-going options.

Formally, we assume the existence of a centralized policy, denoted by $\pi(o_t|z_t)$, which takes the joint observation $z_t$ of all agents as input, and outputs a probabilistic distribution of joint option $o_t$. For mathematical convenience, we introduce a new notation, $U_t$, to denote the set of agents that need to choose new options at option step $t$, and $\overline{U_t}$ to denote the set of agents that do not need to choose new options. The corresponding joint option for $U_t$ and $\overline{U_t}$ are denoted as $o^{U_t}_{t}$ and $o^{\overline{U_t}}_{t}$ respectively.


It turns out that we can take advantage of the concept of \textit{conditional dependency} to derive the policy distribution we need.
More specifically, we formalize a joint policy $\pi(o_t|z_t)$ in an equivalent way of conditioning the option distribution of the agents in $U_t$ based on the currently-executing options of agents in $\overline{U_t}$, as is shown in Eq.~\eqref{eq:fully-cen-1}.
\begin{equation}
\small
\pi(o_t \mid z_t) = \pi(o^{U_t}_{t}, o^{\overline{U_t}}_{t} \mid z_t) \triangleq \hat{\pi}(o^{U_t}_{t} \mid o^{\overline{U_t}}_{t}, z_{t})
\label{eq:fully-cen-1}
\end{equation}
In Eq.~\eqref{eq:fully-cen-1}, we denote $\hat{\pi}$ as a conditional policy to distinguish it from joint policy ${\pi}$, but they shared same parameters.  The use of $\hat{\pi}(o^{U_k}_{k} \mid o^{\overline{U_k}}_{k}, z_k)$ allows us to sample options only for those agents who need to select new options while maintain uninterrupted option executions for the others. Thus, it allows the generation of ``true'' asynchronous trajectory $\tau_o$. Moreover, the generated trajectories $\tau_o$ will be further used as training data to update the policy itself, we thus achieve the on-policy update.

\subsection{Algorithm}
\label{methodC}
In practice, Eq.~\eqref{eq:fully-cen-1} is computationally  feasible. The term $\hat{\pi}(o^{U_t}_{t} \mid o^{\overline{U_t}}_{t}, z_t)$ is a conditional distribution thus can be reformed via Bayes' rule by a joint distribution $\pi(o^{U_t}_{t}, o^{\overline{U_t}}_{t} \mid z_{t})$  and 
its marginalization $\pi(o^{\overline{U_t}}_{t} \mid z_{t})$, as shown in Eq.~\eqref{eq:fully-cen-2}.
\begin{equation}
\small
    o^{U_t}_{t} \sim \hat{\pi}(o^{U_t}_{t} \mid o^{\overline{U_t}}_{t}, z_{t})=\frac{\pi(o^{U_t}_{t}, o^{\overline{U_t}}_{t} \mid z_{t})}{\pi(o^{\overline{U_t}}_{t} \mid z_{t})}
\label{eq:fully-cen-2}
\end{equation}
The joint and marginalized terms can be effortlessly represented and obtained. In many deep RL cases, neural networks are often used to represent policy and value functions. The joint policy $\pi_\theta(o^{U_t}_{t},o^{\overline{U_t}}_{t} \mid z_{t})$ can be represented by a neural network parameterized by $\theta$,  and its output usually stands for the main parameters of commonly-used probabilistic distributions such as categorical or Gaussian distribution.  From these standard distributions, the marginalization $\pi_\theta(o^{\overline{U_t}}_{t} \mid z_{t})$ can also be efficiently computed. For example, marginalization of categorical distribution can be done by summing out the variables corresponding to the categorical distribution to obtain a new distribution over the remaining variables. This process is illustrated in Fig.~\ref{fig:method}'s policy stage.

%

%
%
By utilizing $\hat{\pi}_{\theta}(o^{U_t}_{t} \mid o^{\overline{U_t}}_{t}, z_t)$ as policy and running its optimization over option-level joint trajectory $\tau_o$, we can seamlessly alternate standard MAPG method in Eq.~\eqref{eq:gae_actor_critic} to support the centralized training over asynchronous options, as shown in Eq.~\eqref{eq:gae_actor_critic_gai}. Specifically, the centralized critic $V_{\pi_{\theta}}(z_t)$ and TD($\lambda$) return $G_{t}^{\lambda}$ are computed in the same way as usual, but only based on option-level trajectory. We summarize the entire process in Algo.~\ref{algo}.
\begin{equation} 
\label{eq:gae_actor_critic_gai}
\scriptsize
    \nabla_{\theta} J\left(\hat{\pi}_{\theta}\right)=\underset{\tau_o \sim \hat{\pi}_{\theta}}{\mathrm{E}}\left[\sum_{t} \nabla_{\theta} \log \hat{\pi}_{\theta}\left(o_{t} \mid z_{t}\right)\left(G_{t}^{\lambda} - V_{\hat{\pi}_{\theta}}(z_t)\right)\right]
\end{equation} 
\begin{algorithm}
        \caption{\small Asynchronous Option-Based MAPG}
        \label{algo}
        \small
    \begin{algorithmic}[1]
        \STATE Initialize $N$ agents. Initialize a joint policy $\pi_{\theta}(o|z)$ and a value network $V_{\phi}(z)$. Randomly sample initial observation $z_0$.
        \FOR{iteration$= 0, 1, 2, ...$}
        \STATE Training trajectory dataset $\mathcal{\tau}_o = \{\}$.
        \FOR{$t = 0, 1,2,..., T$} 
        \IF{$U_t \neq \varnothing$ (some agents choose new options)}
        \STATE Obtain $o^{U_t}_{t}$ and $o^{\overline{U_t}}_{t}$, and alternate $\pi_\theta(o_t \mid z_t)$ to be $\hat{\pi}_\theta(o^{U_t}_{t} \mid o^{\overline{U_t}}_{t}, z_{t})$ via Eq.~\eqref{eq:fully-cen-1} and \eqref{eq:fully-cen-2}.
        \STATE Sample $o_t$ from $\hat{\pi}_\theta(o^{U_t}_{t} \mid o^{\overline{U_t}}_{t}, z_{t})$ and execute $o_t$.
        \STATE Add $(z_{t}, o_{t}, r_{t})$ to $\tau_o$.
        \ELSE
        \STATE all agents continue their on-going options.
        \ENDIF
    \ENDFOR
    \STATE Update policy parameter $\theta$ via Eq.~\eqref{eq:gae_actor_critic_gai} using data from ${\tau}_o$.
    \ENDFOR
\end{algorithmic}
\end{algorithm}

\subsection{An Extended Use-Case}
\label{sec:partially-cen}
In addition to centralized learning, our method can also be applied to another important use-case called partially centralized learning. 
In contrast to (fully) centralized scenarios where global states and observations are needed to make global decisions, partially centralized learning describes agents choose their options based on their local observations as well as conditioned by other agents' options.

Partially centralized learning is useful in some realistic situations when having a fully central system is impracticable. For example,  broadcasting global information for all agents can sometimes be costly and restricted by software or hardware capability such as communication bandwidth and latency, especially when global information includes high-dimensional observations (e.g. images or point cloud). 
However, by only sharing options instead of observations, partially centralized learning can be more efficient since the option space is usually low-dimensional. In addition, knowing options of other agents as conditions is potentially important for local decision-making.

Formally, we have multiple policies $\pi^i$, one for each agent $i$ for option selection. Similar to centralized learning, at each option step $t$, we formulate the option selection probability of each agent in $U_t$ by conditioning it on its local observation $z_t^i$ as well as  the currently executing options of agents in $\overline{U_t}$. 
%
%
For agents in the set $\overline{U_t}$, we can simply evaluate its option selection probability to be deterministic and equal to $1$. This is because the agent does not select a new option but naturally continue the on-going option, thus does not contribute to the policy gradient at this option step.
This is  shown in Eq.~\eqref{eq:partial-cen}. 
\begin{equation}
\label{eq:partial-cen}
\small
\hat{\pi}^{i}(o^{i}_{t} \mid o^{\overline{U_t}}_{t}, z^{i}_{t})=\left\{\begin{array}{cl}
1, & \text { if } {i} \in \overline{{U_t}} \\
\frac{\pi(o^{i}_{t}, o^{\overline{U_t}}_{t} \mid z^{i}_{t})}{\pi_(o^{\overline{U_t}}_{t} \mid z^{i}_{t})}, & \text { otherwise. }
\end{array}\right.
\end{equation}

\begin{figure}[!htp]
\centering
\begin{subfigure}[c]{0.55\columnwidth}
\centering
\includegraphics[width=\textwidth]{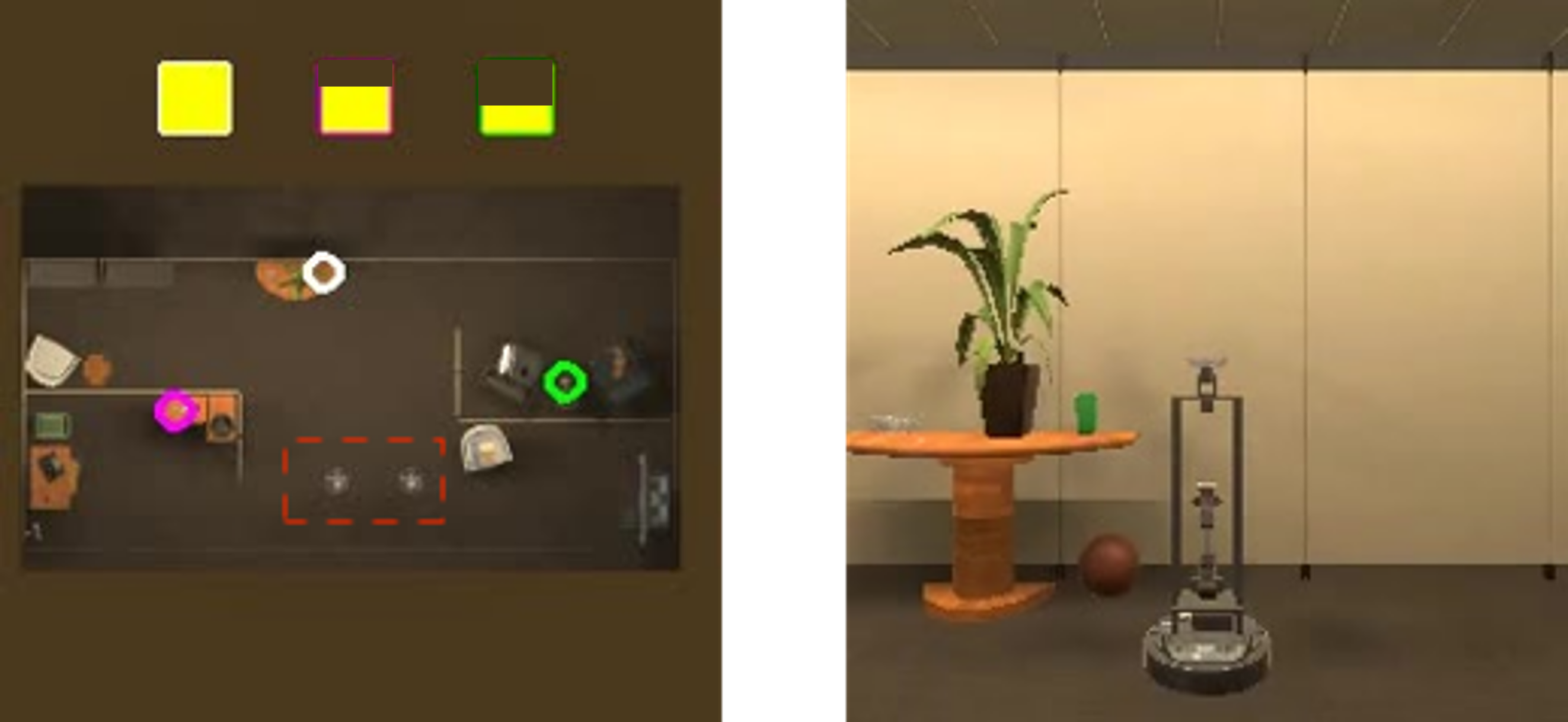}
\caption{}
\label{fig:water fill}
\end{subfigure}%
\hfill
\begin{subfigure}[c]{0.45\columnwidth}
\centering
\includegraphics[width=0.8\textwidth]{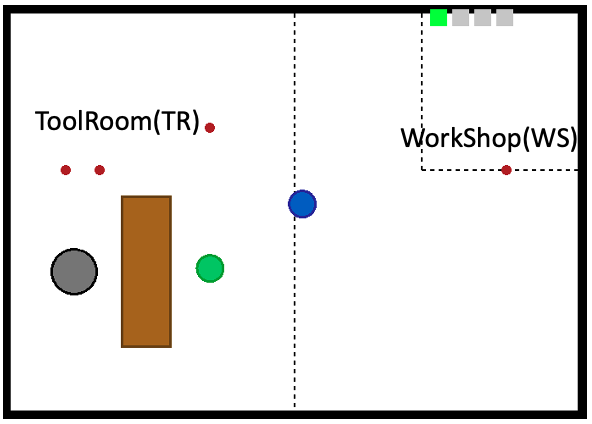}
\caption{}
\label{fig:tool delivery}
\end{subfigure}%
\hfill
\vspace{-6pt}
\caption{\small (a) Water Filling task, top and front view. Water levels are in yellow. Positions of three jars are circled in white, pink and green. Two robots are in the middle space (red dashed box). (b) Tool Delivery task with one Fetchbot (gray) and two Turtlebots (green and blue). Red dots are available middle way-points for transitions, and brown rectangle is desk for passing tools. Progress bar on the top right represents the current task stage.}
\label{fig:task_config}
\end{figure}

\section{Experimental Results}

\subsection{Task Specifications}
We empirically evaluate our method on two cooperative multi-robot tasks featuring in asynchronous options and long-term reasoning, following previous work \cite{xiao2022asynchronous, pmlr-v100-xiao20a}.

\textbf{Water Filling (WF).} 
As shown in Fig.~\ref{fig:water fill}, two heterogeneous robots (a slower vehicle and a faster drone) navigate in a large indoor room and aim to cooperatively refill water timely for three jars at different locations. This task is  challenging as (1) the water levels in three jars decrease randomly and rapidly following parameter-varying Gaussian distributions at each low-level step $k$; (2) two robots have different but limited abilities: vehicle can scout and fill water but move slowly; drone is faster but can only scout water levels. So,  robots have to develop ``real" cooperation to 
make use of their abilities rather than repeating  rote behaviors. For example, drone is expected to scout the room quickly and share the most recent water levels with vehicle. Then vehicle can choose closest or empty jar to fill.

We use ai2thor simulator \cite{kolve2017ai2} to model this task and use RGB raw images as well as important auxiliary information\footnote{In particular, we use (a) global water levels recently updated by any robot; (b) elapsed time steps since water levels get updated last time. } as observations. Both agents can take one-step options: \{up; down; left; right\} or multi-step options: NavTo($c$) where $c$ is jar index. Vehicle can take an extra option NavToFill($c$) to reach and fill water. The low-level time steps of NavTo($c$) and NavToFill($c$) depends on the robot's speed and its distance to target jar: we use $0.5$m/s for vehicle and $2$m/s for drone. Reward function is defined over the global water level of all jars in Eq.~\eqref{eq:wf_rew} where $w(s;c)$ is the water level of jar indexed by $c$ under true state $s$.
\begin{equation}
\label{eq:wf_rew}
\small
r(s) = \sum_{c=0}^{|C|}(-\frac{1.0}{w(s;c)+0.001}+1),\mathrm{c}=0,1,2
\end{equation}

\textbf{Tool Delivery (TD).} 
To validate the usefulness of our method on more complex problems, we use Tool Delivery \cite{pmlr-v100-xiao20a}, a three-agents task that requires
complicated multi-stage collaboration, as shown in Figure.~\ref{fig:tool delivery}. It involves a human working on a four-stage task, requiring assistance from a Fetchbot and two Turtlebots to search, pass and deliver proper tools to him at proper stage. Only Fetchbot can search and pass tools to Turtlebots, and only Turtlebots can deliver tools to human. Turtlebots can choose multi-step options from \{GoToWS; GoToTR; GetTool\} to go to one of middle way-points (TR or WS, red dots in Fig.~\ref{fig:tool delivery}) or pick one tool. Fetchbot either search specific tool or pass one of the found tools to one of the waiting Turtlebots. More constraints of this task can be found in \cite{pmlr-v100-xiao20a}.

Turtlebot observes its own location, human working stage, index of carried tools and the number of the tools on the desk.  Fetchbot observes the number of tools waiting to be passed and the index of waiting Turtlebot. Note that neither Fetchbot nor Turtlebots is aware of the correct tool required by human at each stage, they have to reason about this information via training. The agents receive $-1$ reward at each  low-level step, $-10$ when the Fetchbot passes a tool but no Turtlebot around, and $100$ for a good tool delivery to human.
Available options for all tasks are shown in Table~\ref{tab:option table}.

\begin{table}[ht]
\scriptsize
\centering
\begin{tabulary}{0.5\linewidth}{llp{6cm}}
\toprule
\hfil Task & \hfil Agent & \hfil Available Options  \hfil \\
\midrule
\centering
\hfil WF & \hfil Drone & \hfil Up; Down; Left; Right; NavTo(c); \hfil \\
\hfil WF & \hfil Vehicle & \hfil  Up; Down; Left; Right; Fill(c); NavTo(c); NavToFill(c); \hfil \\
\hfil TD & \hfil Turtlebot & \hfil GoToWS; GoToTR; GetTool  \\
\hfil TD & \hfil Fetchbot & \hfil  SearchTool($j$); PassTo($w$)  \\
\bottomrule
\end{tabulary}
\caption{\small Available agent options for two tasks. Options are discrete. $c$ is the index of jars from WF task. $j \in $ \{0,1,2\} is the tool index and $w \in $ \{0,1\} is the index of waiting Turtlebot from TD task.  Options  have varying low-level time lengths depending on their termination conditions and properties.}
\label{tab:option table}
\vspace{-10pt}
\end{table}%

\subsection{Implementation and Baselines}
This section outlines the methods employed in our study, which includes our proposed approach, baselines, and a state-of-the-art method. In all methods, a joint policy network is used for centralized policy learning across all agents.

Our method and baselines are both adapted from MAPPO \cite{yu2021surprising}, a well-known, action-based MAPG algorithm. 
Our method alters MAPPO to allow the use of a conditional centralized policy and use it to sample asynchronous, option-level joint trajectories for training. 
Baseline methods also use option-level trajectories as training data, but the trajectories are collected synchronously.  They apply use synchronous strategies  such as $\eta_{any}$ and $\eta_{all}$  to forcibly select options for all agents, and collect $\{(z_t, o_t, r_t) \mid t=0,1,...,T\}$ at every option step.  We denote these methods as \textit{``sync-cut''} and \textit{``sync-wait''} respectively. 

We include an \textit{``end2end''} baseline which uses MAPPO to learn a centralized policy purely over low-level actions rather options.
Additionally, we include a state-of-the-art approach that is also applicable to ``interruption-free'', asynchronous option-based data, but mainly uses Q-value based off-policy algorithms. We use mean reward over a batch of training samples per iteration as performance criteria.

\begin{figure*}[!htp]
\centering
\begin{subfigure}[c]{0.5\linewidth}
\centering
\includegraphics[width=0.9\textwidth]{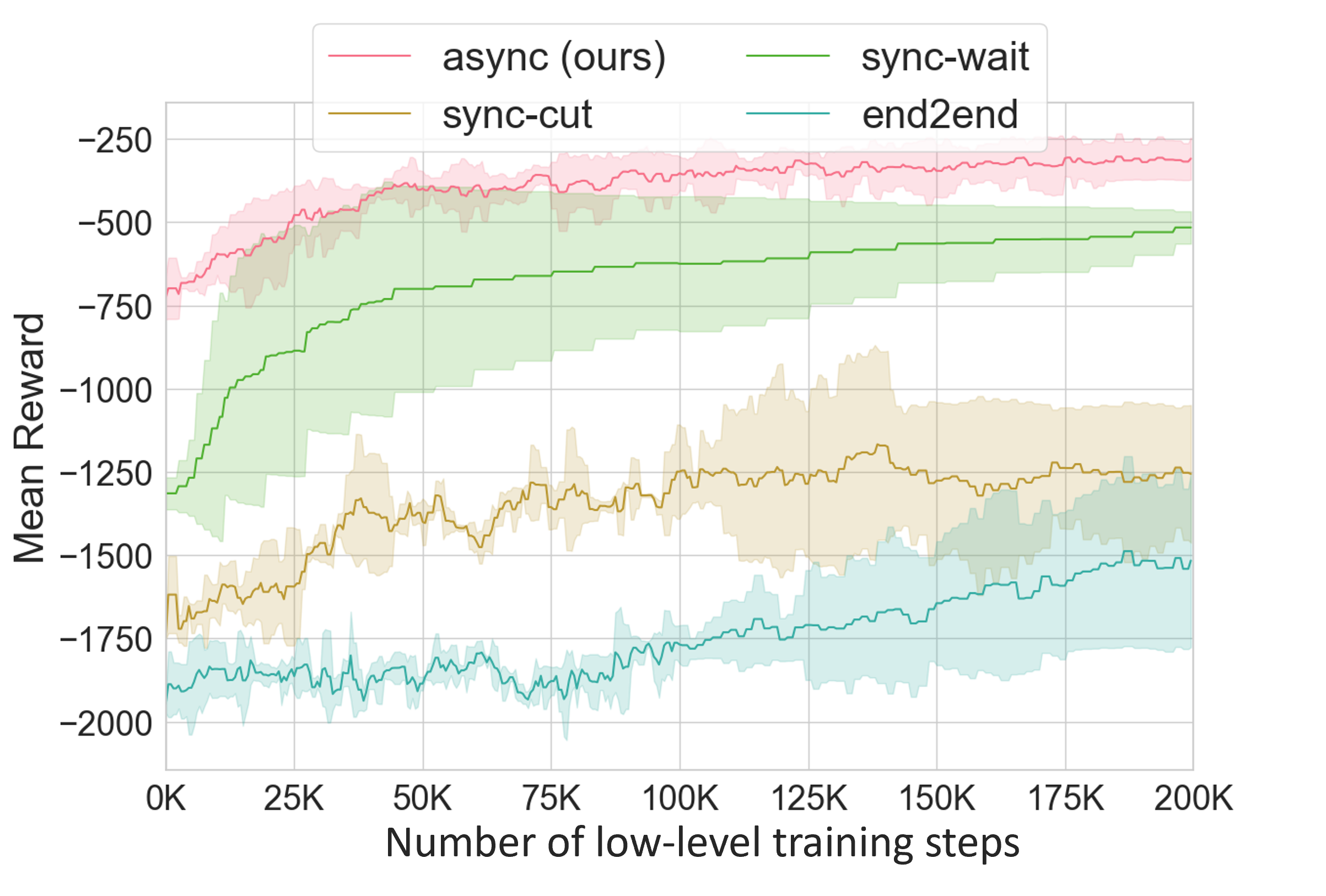}
\captionsetup{justification=centering}
\caption{}
\label{fig:wf_sync_e2e}
\end{subfigure}%
\hfill
\begin{subfigure}[c]{0.5\linewidth}
\centering
\includegraphics[width=0.8\textwidth]{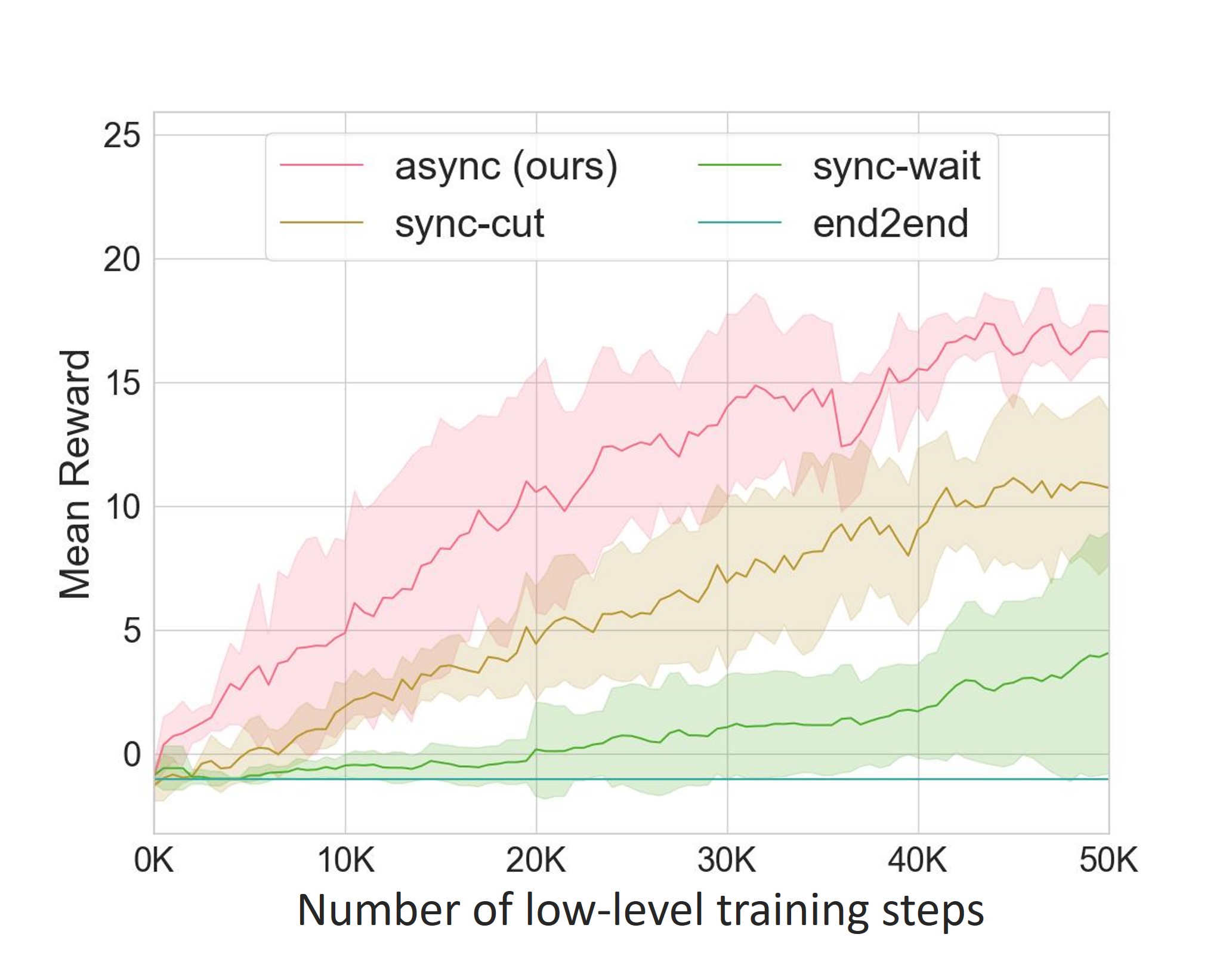} 
\captionsetup{justification=centering}
\caption{}
\label{fig:async-sync}
\end{subfigure}%
\hfill
\vspace{-6pt}
\caption{\small  Centralized policy learning using asynchronous option-based (ours), synchronous option-based as well as low-level action-based methods on {(a)} Water Filling and {(b)} Tool Delivery task. 
All experimental results are summarized over 5 runs with random seeds.
%
During each training epoch, we gather a fixed number of low-level samples, and extract option-based trajectory data from them to update the policy.
The variation of starting reward levels in (a) is attributed to the use of different option termination strategies, which can impact the number of option steps included in a trajectory. Thus, the variation in trajectory length influences the cumulative reward attained.
}
\label{res:async-sync}
\end{figure*}

For all methods, we set the training batch size to be based on low-level action step instead of high-level option step. During each training epoch,  a fixed batch size of low-level samples is collected. For the WF task, batch size is 2000 while for the TD task, batch size is 200. Despite the variation in the number of option steps in a batch due to different option termination strategies. we ensure a fair comparison of ``convergence rate v.s. sample size'' (Fig.~\ref{res:async-sync}) between different methods, particularly between option-based and action-based methods, as an option often contains more than one action.

\subsection{Performance Comparison with Baselines}
%
Fig.~\ref{fig:wf_sync_e2e} shows the actual training progress of ours and baseline methods on Water Filling task. Our approach, denoted by \textit{``async(ours)''},  achieves consistently increasing mean reward, and outperforms other baselines from starting to the end. This indicates the effectiveness of our centralized, conditional-based policy gradient on using complete asynchronous options, which is beneficial to long-term planning, especially in the large indoor environment, otherwise the interrupted options (e.g. using \textit{``sync-cut''}) often give rise to navigation failure in the middle of path and waiting other agents' options can lead to very inefficient exploration process (e.g. using \textit{``sync-wait''}, drone has to wait for slower vehicle and cannot scout the environment information quickly and frequently).

In Fig.~\ref{fig:async-sync}, we show the training results on Tool Delivery task.  Clearly, our approach still acquires the highest peak reward as well as the fastest convergence rate in contrast to the other baselines, despite  more agents, stages as well as cooperative structures involved. This again indicates advantage of being capable of running MAPG on complete, asynchronous options over on forced synchronized options. What even worse is, without option-level operation, ``end2end'' method cannot learn any useful policy at all.

\subsection{Performance Comparison with a state-of-the-art method}
%
To show the effectiveness of our method is not only stemmed from running complete options but also the method itself,  we present  Table.~\ref{tab:sota table}, which summarizes the results of final policy trained by ours and a state-of-the-art method ``Mac-DQN'' \cite{pmlr-v100-xiao20a}, in terms of its peak performance and sample efficiency. For both methods, we use mean reward as a consistent performance criteria. It indicates that despite both methods use ``true'' asynchronous option without any forced synchronization, ours still achieves not only a comparable level of peak performance, but also much better sample efficiency than ``Mac-DQN'' on both tasks (3$\sim$8x faster).

\begin{table}[!htp]
\centering
\small
\setlength{\tabcolsep}{5pt}
\begin{tabular*}{\columnwidth}{ccccc}
\toprule
              & \multicolumn{2}{c}{{Peak Performance}} & \multicolumn{2}{c}{{Training Samples}} \\
\midrule
              & {Mac-DQN}            & {Ours}            &{Mac-DQN}            & {Ours}            \\
{Water Filling} & {$-287 \pm 41$ }              & {$\bm{-304 \pm 27}$}            & {$600$K}               & {$\bm{200}$K}            \\
{Tool Delivery} & {$14.2 \pm 0.8$}               & {$\bm{17.1 \pm 0.7}$}           & {$400$K}               & {$\bm{50}$K}          \\
\bottomrule
\end{tabular*}
\caption{\small Performance comparison with Mac-DQN, a state-of-the-art asynchronous option-based MAPG method using Q-value based, off-policy optimization.}
\label{tab:sota table}
\vspace{-8pt}
\end{table}

\subsection{Centralized Learning: Fully v.s. Partially}

We evaluate performance of policies  trained by fully and partially centralized learning, as shown in Table.~\ref{tab:partially centralized table}. It was found that partially centralized learning achieved relatively close performance to centralized learning in the Water Filling task, but performed significantly worse in the Tool Delivery task. This may be because in the Water Filling task, important task states such as the global water level can be effectively communicated between robots and encoded in their local observations, and thus having only local observations is sufficient for robots to navigate and search effectively. However, in the multi-stage Tool Delivery task, more complex cooperation and global observation sharing is necessary to obtain important task states for all agents. Therefore, using partially centralized learning is not sufficient for making good decisions based solely on local observations. Overall, partially centralized learning may be useful in cases where observation sharing is less critical but option-dependency is more important, as it can reduce computational load.

\begin{table}[!htp]
\centering
\small
\begin{tabular}{ccc}

\toprule
              & {Partially Centralized} & {Centralized (Ours)} \\
\midrule              
{Water Filling} & {$-349 \pm 51$}                 & {$\bm{-304 \pm 27}$}        \\
{Tool Delivery} & {$4.4 \pm 0.3$}                   & {$\bm{17.1 \pm 0.7}$}  \\
\bottomrule
\end{tabular}%
\caption{\small Performance comparison between partially centralized and centralized learning. Both use asynchronous options.}
\label{tab:partially centralized table}
\vspace{-12pt}
\end{table}

\section{Conclusion}
We propose a conditional reasoning approach to enable MAPG methods to learn {centralized policies} over asynchronous options while preserving options' asynchronous executions. It seamlessly adapts off-the-shelf MAPG methods from action-based to option-based, and has potentials to be deployed for realistic multi-robot system where asynchronous cooperation is often needed. Experimental results indicate the  effectiveness of our method on two complex
multi-agent cooperative tasks through empirical validation.

\bibliography{main}
\bibliographystyle{IEEEtran.bst}
\end{document}